\documentclass[11pt]{article}
\usepackage{amsfonts, amssymb, amsmath}
\usepackage{mtsummit2019}
\usepackage{times}
\usepackage{url}
\usepackage{latexsym}
\usepackage{graphicx}
\usepackage{subcaption}
\usepackage[export]{adjustbox}
\usepackage{float}
\usepackage{caption} 

\newcommand{\norm}[1]{\left\lVert#1\right\rVert}

\title{Memory-Augmented Neural Networks for Machine Translation}

\author{Mark Collier\\
  Trinity College Dublin\\
  School of Computer Science and\\
  Statistics, Dublin, Ireland\\
  {\tt mcollier@tcd.ie}  \And
  Joeran Beel\\
Trinity College Dublin\\
  School of Computer Science and \\
  Statistics, ADAPT Centre, Dublin, Ireland\\
  {\tt joeran.beel@tcd.ie}}

\date{}

\begin{document}
\maketitle
\begin{abstract}
  Memory-augmented neural networks (MANNs) have been shown to outperform other recurrent neural network architectures on a series of artificial sequence learning tasks, yet they have had limited application to real-world tasks. We evaluate direct application of Neural Turing Machines (NTM) and Differentiable Neural Computers (DNC) to machine translation. We further propose and evaluate two models which extend the attentional encoder-decoder with capabilities inspired by memory augmented neural networks. We evaluate our proposed models on IWSLT Vietnamese$\rightarrow$English and ACL Romanian$\rightarrow$English datasets. Our proposed models and the memory augmented neural networks perform similarly to the attentional encoder-decoder on the Vietnamese$\rightarrow$English translation task while have a 0.3-1.9 lower BLEU score for the Romanian$\rightarrow$English task. Interestingly, our analysis shows that despite being equipped with additional flexibility and being randomly initialized memory augmented neural networks learn an algorithm for machine translation almost identical to the attentional encoder-decoder.
\end{abstract}

\section{Introduction}

Memory-Augmented Neural Networks (\textbf{MANN}) are a new class of recurrent neural network (\textbf{RNN}) that separate computation from memory. The key distinction between MANNs and other RNNs such as Long Short-Term Memory cells (\textbf{LSTM}) \cite{RN18} is the existence of an external memory unit. A controller network in the MANN receives input, interacts with the external memory unit via read and write heads and produces output. MANNs have been shown to learn faster and generalize better than LSTMs on a range of artificial sequential learning tasks \cite{RN11,RN12,RN38}. Despite their success on artificial tasks, LSTM based models remain the preferred choice for many commercially important sequence learning tasks such as handwriting recognition \cite{RN36}, machine translation \cite{RN16} and speech recognition \cite{graves2014towards}.

Attentional encoder-decoders \cite{RN6,RN10}, where the encoder and decoder are often LSTMs or other gated RNNs such as the Gated Recurrent Unit \cite{cho2014learning}, are a class of neural network models that have achieved state-of-the-art performance on many language pairs for machine translation \cite{RN49,RN60}. An encoder RNN reads the source sentence one token at a time. The encoder both maintains an internal vector representing the full source sentence and it encodes each token in the source sentence into a vector often assumed to represent the meaning of that token in its surrounding context. The decoder receives the internal vector from the encoder and can read from the encoded source sentence when producing the target sentence.

Attentional encoder-decoders can be seen as a basic form of MANN. The collection of vectors representing the encoded source sentence can be viewed as external memory which is written to by the encoder and read from by the decoder. But attentional encoder-decoders do not have the same range of capabilities as MANNs such as the Neural Turing Machine (\textbf{NTM}) \cite{RN11} or Differentiable Neural Computer (\textbf{DNC}) \cite{RN12}. The encoder RNN in attentional encoder-decoders must write a vector at each timestep and this write must be to a single memory location. The encoder is not able to update previously written vectors and has only one write head. The decoder has read only access to the encoded source sentence and typically just a single read head. Widely used attention mechanisms \cite{RN6,RN10} do not have the ability to iterate through the source sentence from a previously attended location. All of these capabilities are present in NTMs and DNCs.

In this paper we propose two extensions to the attentional encoder-decoder which add several capabilities present in other MANNs. We are also the first that we are aware of to evaluate the performance of MANNs applied directly to machine translation.

\section{Background}

We briefly review how attention weights are computed for Luong attention \cite{RN10} and how addresses are computed for the NTM. Alternative attention mechanisms have similar computations \cite{RN6} and likewise for alternative MANNs such as DNCs \cite{RN12}.

\subsection{Luong Attention}

At each timestep $t$ during the decoding of an attentional encoder-decoder a weighting $\mathbf{w_t}$ over the encoded source sentence is computed, where $\sum_s w_t(s) = 1$ and $\forall s \ w_t(s) \geq 0$. The predicted token at that timestep during decoding is then a function of the decoder RNN hidden state $\mathbf{h}_t$ and the weighted sum of the encoder hidden states i.e. $\sum_{s} w_t (s) * \mathbf{\hat{h}}_{s}$.

The difference between various attention mechanisms is how to compute the weighting $\mathbf{w_t}$. In Luong attention \cite{RN10} the weighting is computed as the softmax over scaled scores for each source sentence token, eq.\ \ref{eq:content_based_addr_1}. The scores for each source sentence token are computed as the dot product of decoder RNN hidden state $\mathbf{h}_t$ and encoder RNN hidden state $\mathbf{\hat{h}}_s$ which is first linearly transformed by a matrix $\mathbf{W}_a$.

\begin{equation} \label{eq:loung_score_1}
score(\mathbf{h}_t, \mathbf{\hat{h}}_s) \gets \mathbf{h}_{t}^{\top} \mathbf{W}_a \mathbf{\hat{h}}_s
\end{equation}

\begin{equation} \label{eq:content_based_addr_1}
w_t(s) \gets \frac{\exp(\beta_t * score(\mathbf{h}_t, \mathbf{\hat{h}}_s))}{\sum_{s'} \exp(\beta_t * score(\mathbf{h}_t, \mathbf{\hat{h}}_{s'}))}
\end{equation}

\subsection{NTM Addressing}

Rather than computing weightings over an encoded source sentence, NTMs have a fixed sized external memory unit which is a $N * W$ memory matrix. $N$ represents the number of memory locations and $W$ the dimension of each memory cell. A controller neural network has read and write heads into the memory matrix. Addresses for read and write heads in a NTM are computed somewhat similarly to attention mechanisms. However in addition to being able to address memory using the similarity between a lookup key and memory contents, so called content based addressing, NTMs also have the ability to iterate from current or past addresses. This enables NTMs to learn a broader class of algorithms than attentional encoder-decoders \cite{RN11,RN12}.

At each timestep ($t$), for each read and write head the controller network outputs a set of parameters; a lookup key $\mathbf{k}_t$, a scaling factor $\beta_t \geq 0$, an interpolation gate $g_t \in [0, 1]$,  a shift kernel $\mathbf{s}_t$ (s.t. $\sum_k s_t(k) = 1$ and $\forall k \ s_t(k) \geq 0$) and a sharpening parameter $\gamma_t \geq 1$ which are used to compute the weighting $\mathbf{w}_t$ over the N memory locations in the memory matrix $\mathbf{M}_t$ as follows:

\begin{equation} \label{eq:content_based_addr_2}
w^c_t(i) \gets \frac{\exp(\beta_t * K[\mathbf{k}_t, \mathbf{M}_t(i)])}{\sum_{j=0}^{N-1} \exp(\beta_t * K[\mathbf{k}_t, \mathbf{M}_t(j)])}
\end{equation}

We can see that $\mathbf{w}^c_t$ is computed similarly to Luong attention and allows for content based addressing. $\mathbf{k}_t$ represents a lookup key into memory and $K$ is some similarity measure such as cosine similarity:

\begin{equation} \label{eq:cosine_similarity}
K[\mathbf{u}, \mathbf{v}] = \frac{\mathbf{u} \cdot \mathbf{v}}{\norm{\mathbf{u}} \cdot \norm{\mathbf{v}}}
\end{equation}

NTMs enable iteration from current or previously computed memory weights as follows:

\begin{equation} \label{eq:interpolation_1}
\mathbf{w}^{g}_{t} \gets g_t \mathbf{w}^c_t + (1 - g_t) \mathbf{w}_{t-1}
\end{equation}

\begin{equation} \label{eq:conv_shift_eq_1}
\tilde{w}_{t}(i) \gets \sum_{j=0}^{N-1} w^{g}_{t}(j) s_t(i-j)
\end{equation}

\begin{equation} \label{eq:sharpening_1}
w_{t}(i) \gets \frac{\tilde{w}_{t}(i)^{\gamma_t}}{\sum_{j=0}^{N-1} \tilde{w}_{t}(j)^{\gamma_t}}
\end{equation}

\noindent where (\ref{eq:interpolation_1}) enables the network to choose whether to use the current content based weights or the previous weight vector, (\ref{eq:conv_shift_eq_1}) enables iteration through memory by convolving the current weighting by a 1-D convolutional shift kernel and (\ref{eq:sharpening_1}) corrects for any blurring occurring as a result of the convolution operation.

The vector $\mathbf{r}_t$ read by a particular read head at timestep $t$ is computed as a weighted sum over memory locations similarly to Luong attention:

\begin{equation} \label{eq:read}
\mathbf{r}_t \gets \sum_{i=0}^{N-1} w_t(i) *  \mathbf{M}_t(i)
\end{equation}

An attentional encoder-decoder has no write mechanism. Another way to view this, is that an attentional encoder-decoder has a memory matrix with $N$ equal to the source sentence length and the encoder must always write its hidden state to the memory location corresponding to its position in the source sentence. A NTM does have a write operation, with write addresses determining a weighting over memory locations for the write. Each write head modifies the memory matrix by outputting erase ($\mathbf{e}_t$) and add ($\mathbf{a}_t$) vectors which are then used to softly zero out existing memory contents and write new memory contents through addition:

\begin{equation} \label{eq:write_erase}
\tilde{\mathbf{M}}_t(i) \gets \mathbf{M}_{t-1}(i)[\mathbf{1} - w_t(i) \mathbf{e}_t]
\end{equation}

\begin{equation} \label{eq:write_add}
\mathbf{M}_t(i) \gets \tilde{\mathbf{M}}_t(i) + w_t(i) \mathbf{a}_t
\end{equation}

\section{Proposed Models}

We propose two models which bridge the gap between the attentional encoder-decoder and MANNs, extending the attentional encoder-decoder with additional mechanisms inspired by MANNs. We also propose the application of MANNs directly to machine translation.

\subsection{Neural Turing Machine Style Attention}

The reads from a decoder in an attentional encoder-decoder for machine translation often exhibit monotonic iteration through the encoded source sentence \cite{RN6,raffel2017online}. However widely used attention mechanisms have no way to explicitly encode such a strategy. NTMs combine content based addressing similar to attention mechanisms with the ability to iterate through memory. We propose a new attention mechanism which combines the content based addressing of Luong attention \cite{RN10} with the ability to iterate through memory from NTMs. For our proposed attention mechanism at each timestep ($t$) the decoder outputs a set of parameters for each of its read heads: $\mathbf{h}_t$, $\beta_t \geq 0$, $g_t \in [0, 1]$, $\mathbf{s}_t$ (s.t. $\sum_k s_t(k) = 1$ and $\forall k \ s_t(k) \geq 0$) and $\gamma_t \geq 1$ which are used to compute the weighting $\mathbf{w}_t$ over encoded source sentence $\mathbf{\hat{h}}_s$ for $s = 1,2,...$

\begin{equation} \label{eq:loung_score}
score(\mathbf{h}_t, \mathbf{\hat{h}}_s) \gets \mathbf{h}_{t}^{\top} \mathbf{W}_a \mathbf{\hat{h}}_s
\end{equation}

\begin{equation} \label{eq:content_based_addr}
w^c_t(s) \gets \frac{\exp(\beta_t * score(\mathbf{h}_t, \mathbf{\hat{h}}_s))}{\sum_{s'} \exp(\beta_t * score(\mathbf{h}_t, \mathbf{\hat{h}}_{s'}))}
\end{equation}

\begin{equation} \label{eq:interpolation}
\mathbf{w}^{g}_{t} \gets g_t \mathbf{w}^c_t + (1 - g_t) \mathbf{w}_{t-1}
\end{equation}

\begin{equation} \label{eq:conv_shift_eq}
\tilde{w}_{t}(s) \gets \sum_{j=0}^{N-1} w^{g}_{t}(j) s_t(s-j)
\end{equation}

\begin{equation} \label{eq:sharpening}
w_{t}(s) \gets \frac{\tilde{w}_{t}(s)^{\gamma_t}}{\sum_{j=0}^{N-1} \tilde{w}_{t}(j)^{\gamma_t}}
\end{equation}

Equations (\ref{eq:loung_score}) and (\ref{eq:content_based_addr}) represent the standard content based addressing of Luong style attention. Equations (\ref{eq:interpolation}-\ref{eq:sharpening})  replicate equations (\ref{eq:interpolation_1}-\ref{eq:sharpening_1}) of the NTM to enable iteration from the currently attended source sentence token $\mathbf{w}^c_t$ or the previously attended token $\mathbf{w}_{t-1}$. As with the NTM equation (\ref{eq:conv_shift_eq}) represents a 1D convolution on the weighting $\mathbf{w}^{g}_{t}$ with a convolutional shift kernel which is outputted by the decoder to enable iteration. Equation (\ref{eq:sharpening}) corrects for any blurring resulting from the 1D convolution. We can see that such an attention mechanism has the content based addressing capability of Luong attention and the same capability to iterate from previously computed addresses as NTMs.

\subsection{Memory-Augmented Decoder (M.A.D)}

\begin{figure*}[h]
\includegraphics[width=0.99\linewidth]{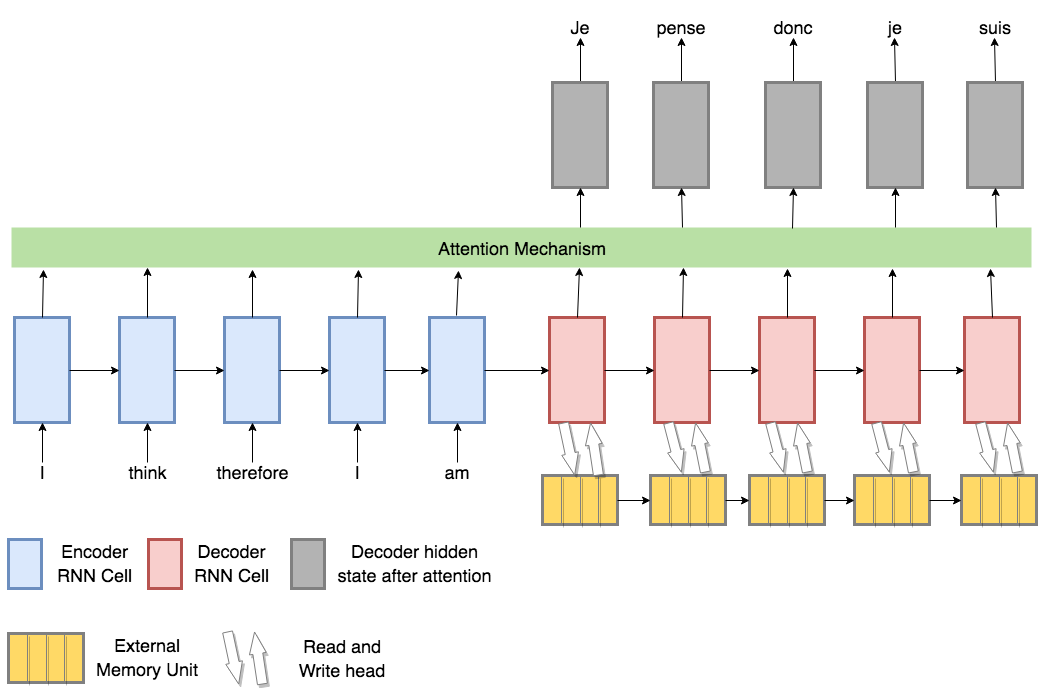} 
\caption{Memory augmented decoder}
\label{fig:mad}
\end{figure*}

The introduction of attention mechanisms has proved highly successful for neural machine translation. Attention extends the writable memory capacity of the encoder in an encoder-decoder model linearly with the length of the source sentence. This avoids the bottleneck of having to encode the whole source sentence meaning into the fixed size vector passed from the encoder to decoder \cite{RN6,RN9}. But the decoder in an attentional encoder-decoder must still maintain a history of its past actions in a fixed size vector. We are motivated by the success of attention which extended the memory capacity of the encoder to propose the addition of an external memory unit to the decoder of an attentional encoder-decoder, hence extending the decoder's memory capacity, fig.\ \ref{fig:mad}. We still maintain a read-only attention mechanism into the encoded source sentence, however the decoder now has the ability to read and write to an external memory unit. We can set the external memory unit to have a number of memory locations greater than the maximum target sentence length in the corpus, thus scaling the decoder's memory capacity with the target sentence length in a similar vain as to how attention scaled the encoder's memory capacity with source sentence length.

We note that a similar model has been proposed before \cite{RN25}, but that in order to train their model the authors propose a pre-training approach based on first training without the external memory unit attached to decoder and then adding it on. This approach restricts the form of possible memory interactions as it must be possible to add the external memory unit while maintaining the pre-trained weights of the attentional encoder-decoder. We simply make the decoder a NTM with the standard read and write heads into an external memory and an additional read head into the encoded source sentence with the addresses on this read head computed in Luong attention style, but other choices for the addressing mechanism are possible, including DNC style addressing. Following a recent stable NTM implementation \cite{collier2018implementing} we do not have any problems training our proposed model.

\subsection{Pure Memory-Augmented Neural Network}

\begin{figure*}[h]
\includegraphics[width=0.99\linewidth]{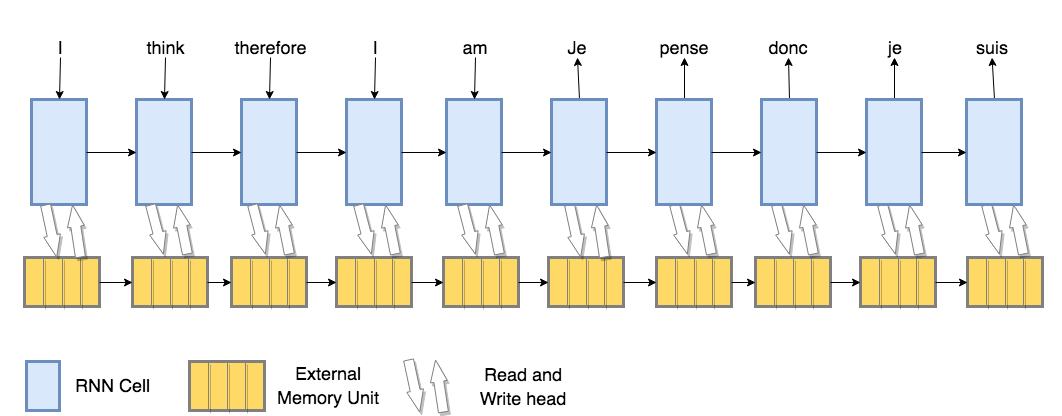} 
\caption{Pure memory augmented neural network for machine translation}
\label{fig:pure_mann}
\end{figure*}

We propose a pure MANN model for machine translation, fig.\ \ref{fig:pure_mann}. Under our proposed model a MANN receives the embedded source sentence as input one token at a time and then receives an end of sequence token. The MANN must then output the target sentence. We are motivated by the enhanced performance of MANNs compared to LSTMs on artificial sequence learning tasks \cite{RN11,RN12,RN59}.

We note that our proposed model has the representational capability to learn a solution similar to an attentional encoder-decoder by simply writing an encoding of each source sentence token to a single memory location and reading the encodings back using content based addressing after the end of sequence token.

We also highlight the differences to the attentional encoder-decoder model. The pure MANN model may have multiple read and write heads each of which uses more powerful addressing mechanisms than popular attention mechanisms. The proposed model may also update previously written locations in light of new information or reuse memory locations if the previous contents have already served their purpose. There is no separation between encoding and decoding and thus only a single RNN is used as the MANN's controller rather than different RNN cells for the encoder and decoder in an encoder-decoder, halving the number of network parameters dedicated to this part of the network.

In this paper for the pure MANN model we use and compare NTMs and DNCs as the choice MANN, however any other MANN with differentiable read and write mechanisms into an external memory unit would be permissible. In both cases we use a LSTM controller. We also compare the use of multiple read and write heads for the NTM model.

\section{Methodology}

We evaluate our models on two machine translation tasks. As a low resource spoken language task we use the 2015 International Workshop on Spoken Language's dataset of English to Vietnamese translated TED talks. We follow  \cite{RN49} in their preprocessing and setup and use their results as a baseline. For training we use TED tst2013, a dataset of 133K  sentence pairs. As the validation set we use TED tst2012 and  test set results are reported on the TED tst2015 dataset. We  use a fixed vocabulary of 17.5K words and 7.7K words for English  and Vietnamese respectively. Any words outside the source or target vocabulary are mapped to an unknown token (UNK).

As a medium resource written language task we follow \cite{RN60} in their general setup for the Romanian to English task from the ACL's 2016 First Conference on Machine Translation's, Machine Translation of News Task. We use their results as a baseline. We train our models on the Europarl English Romanian dataset which consists of 600k sentence pairs. We use the newsdev2016 and newstest2016 datasets as the validation and test sets respectively. We Byte Pair Encode \cite{RN61} the source and target languages with 89,500 merge operations. After Byte  Pair Encoding, the English vocabulary size is 48,824 sub-words and 65,699 sub-words for the Romanian vocabulary.

For all models we use the Adam optimizer \cite{RN15} with an initial learning rate of 0.001. We train for a fixed number of steps but after each epoch we measure the BLEU score on the validation set and measure the test set performance from the version of the model with the highest validation set BLEU score. For the Vietnamese$\rightarrow$English models we train for 14,000 steps and for the Romanian$\rightarrow$English models we train for 120,000 steps.

For all models we use beam search with a beam width of 10. We set the dropout rate to 0.3 with no other regularization applied. For both the Vietnamese$\rightarrow$English and Romanian$\rightarrow$English tasks we follow \cite{RN49,RN60} in using a stack of 2 x 512 unit LSTMs as the encoder and decoder for all relevant models and the controller network for the MANNs. For the memory-augmented decoder the number of memory locations is set to 64 and each memory location is a 512 dimensional vector. Whereas for the pure MANN model the number of memory locations is set to 128 with the memory cell size also set to 512.

We implement our model in Tensorflow, extending Google's NMT implementation \cite{luong17}, and make it available publicly\footnote{\url{https://github.com/MarkPKCollier/MANNs4NMT}}.

\section{Results}

The test set BLEU scores for the Vietnamese$\rightarrow$English translation task are all very similar, with each model's score within the range of 23.1-23.8 BLEU (table \ref{vietnamese_english}). Interestingly, despite the pure MANN models seeing the source sentence in a uni-directional fashion (with all other models using bi-directional encoders) the pure MANN models perform on par with the other models.

\begin{table}[t!]
\begin{center}
\begin{tabular}{|l|c|c|}
\hline \bf Model & \bf Dev & \bf Test \\ \hline
\cite{RN49} & - & 23.3 \\
NTM Style Attention & 21.5 & 23.6 \\
M.A.D. (1 R/W head) & 21.1 & 23.1 \\
M.A.D. (2 R/W heads) & 21.2 & 23.8\\
Pure MANN (NTM - 1 R/W head) & 20.9 & 23.5 \\
Pure MANN (NTM - 2 R/W heads) & 21.3  & 23.5 \\
Pure MANN (DNC - 1 R/W head) & 20.6  & 23.6 \\
\hline
\end{tabular}
\end{center}
\caption{\label{vietnamese_english} Vietnamese$\rightarrow$English translation results (BLEU) on dev (TED tst2012) and test (TED tst2013) sets. M.A.D $\leftrightarrow$ Memory-Augmented Decoder. 1 R/W head means the MANN had 1 read and 1 write head into external memory.}
\end{table}

The attentional encoder-decoder \cite{RN60} has the highest test set BLEU score of all the models for the Romanian$\rightarrow$English translation task (table \ref{romanian_english}). The proposed extensions to the attentional encoder-decoder result in 0.3-0.9 lower test set BLEU score. For the Romanian$\rightarrow$English translation task, the pure MANN model has 1.5-1.9 lower test set BLEU score.

\begin{table}[t!]
\begin{center}
\begin{tabular}{|l|c|c|}
\hline \bf Model & \bf Dev & \bf Test \\ \hline
\cite{RN60} & 30.0 & 29.2 \\
NTM Style Attention & 30.0 & 28.7 \\
M.A.D. (1 R/W head) & 29.8 & 28.9 \\
M.A.D. (2 R/W heads) & 29.7 & 28.3\\
Pure MANN (NTM - 1 R/W head) & 28.9 & 27.7 \\
Pure MANN (NTM - 2 R/W heads) & 28.0  & 27.3 \\
Pure MANN (DNC - 1 R/W head) & 27.8 & 27.5 \\
\hline
\end{tabular}
\end{center}
\caption{\label{romanian_english} Romanian$\rightarrow$English translation results (BLEU) on dev (newsdev2016) and test (newstest2016) sets. M.A.D $\leftrightarrow$ Memory-Augmented Decoder. 1 R/W head means the MANN had 1 read and 1 write head into external memory.}
\end{table}

\subsection{Analysis}

We now examine the attention weights for an attentional encoder-decoder and address computation for the 1 R/W head NTM on a particular Romanian$\rightarrow$English translation. The sentence was chosen as it was the first sentence in our test set which had the same translation from both models. We note that the pattern of addresses are typical of the addresses computed on other sentences for both language pairs, but that a single typical example is presented for brevity.

\begin{figure}[H]
\includegraphics[width=0.99\linewidth]{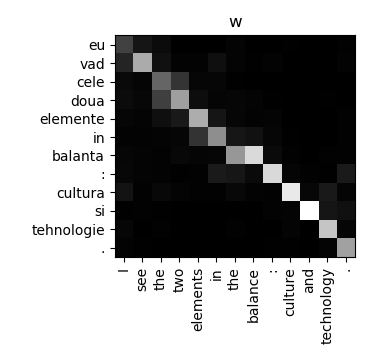} 
\caption{Example attention weights for attentional encoder-decoder}
\label{fig:attention_w}
\end{figure}

We see that we replicate the monotonic iteration through the source sentence often observed in attention mechanisms \cite{RN6,raffel2017online} in fig.\ \ref{fig:attention_w}. We note that this pattern of addressing must be computed solely using the content based addressing of the attention mechanism as no iteration capability is available to the attention weight computation.

\begin{figure}[H]
 
\begin{subfigure}{0.48\textwidth}
\includegraphics[width=0.99\linewidth]{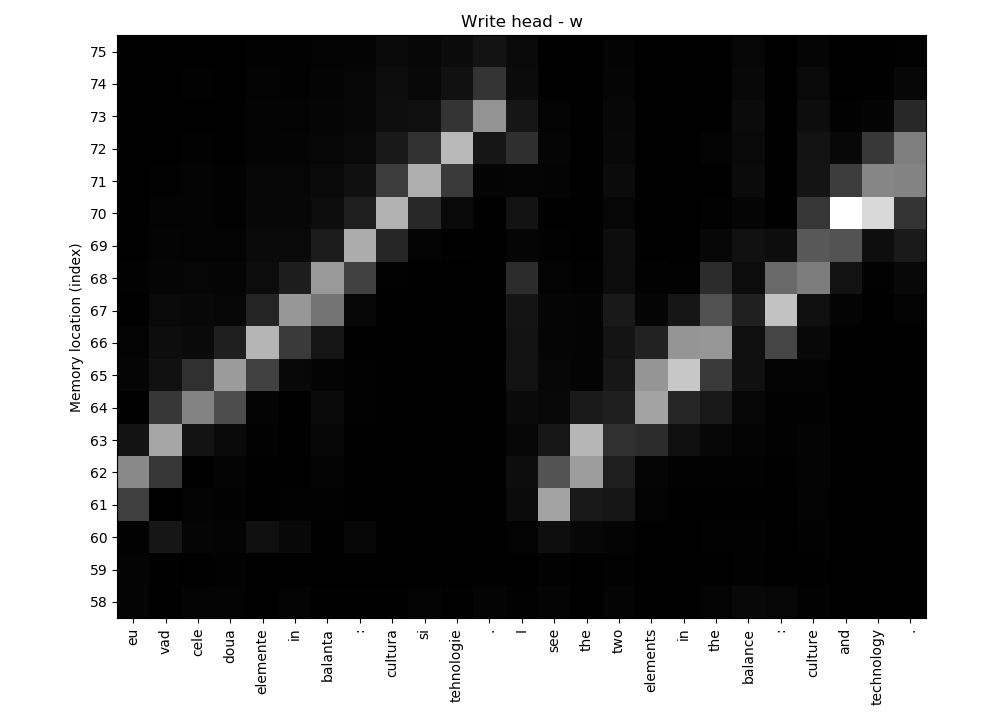} 
\caption{Full address}
\label{fig:write_head_w}
\end{subfigure}

\begin{subfigure}{0.48\textwidth}
\includegraphics[width=0.99\linewidth]{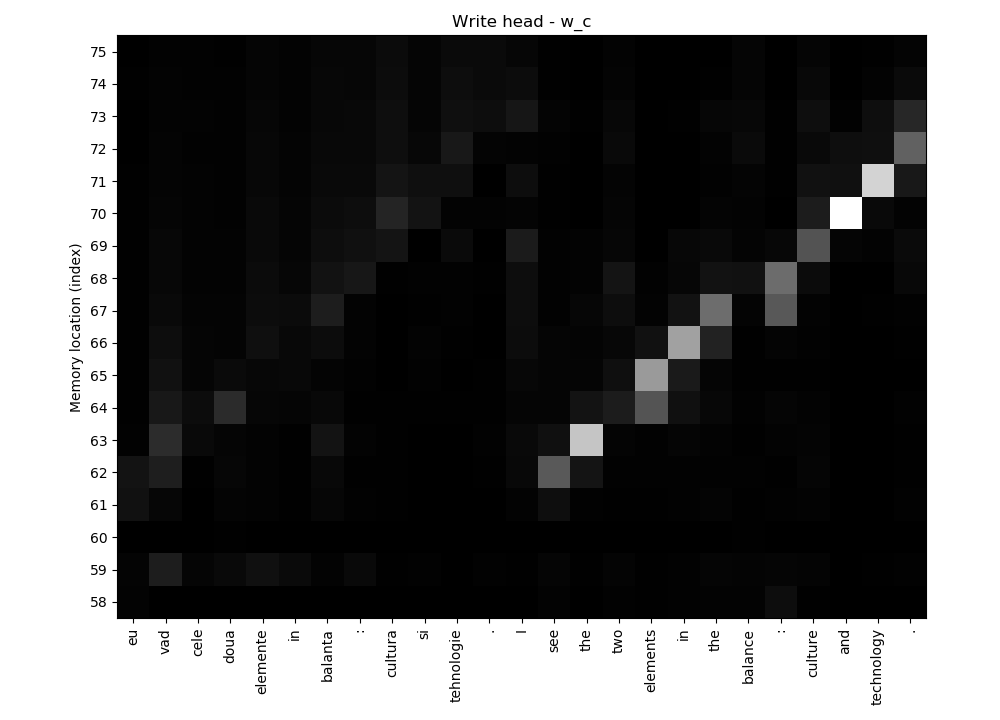}
\caption{Content based addressing}
\label{fig:write_head_w_c}
\end{subfigure}

\begin{subfigure}{0.48\textwidth}
\includegraphics[width=0.99\linewidth]{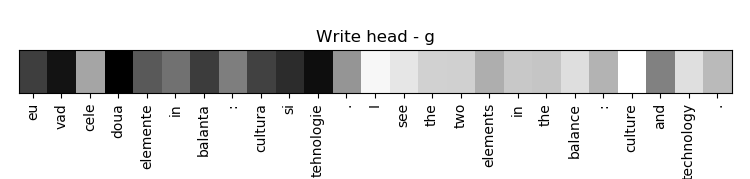}
\caption{Interpolation gate}
\label{fig:write_head_g}
\end{subfigure}

\begin{subfigure}{0.48\textwidth}
\includegraphics[width=0.99\linewidth]{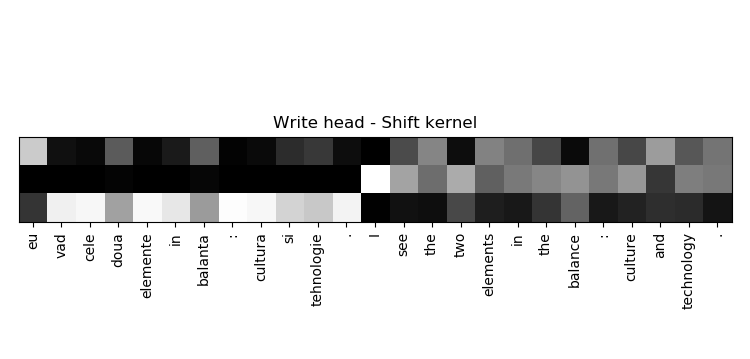}
\caption{Shift kernel}
\label{fig:write_head_s}
\end{subfigure}
 
\caption{Example write head address computation}
\label{fig:write_head}
\end{figure}

We now examine how the NTM has computed the addresses for its read and write head in order to arrive at the same resulting translation. Looking first at the write head, fig.\ \ref{fig:write_head}, we see that as the NTM is shown the source sentence it has learned a very similar strategy to the encoder of an attentional encoder-decoder. In particular we can see that the write head writes an encoded version of the source sentence tokens to successive memory locations, fig.\ \ref{fig:write_head_w}. Interestingly we see that the successive memory locations are computed using the iteration cabability of the NTM as the content based addresses are not significant, fig.\ \ref{fig:write_head_w_c} and the shift kernel is iterating forward through memory, fig.\ \ref{fig:write_head_s} from the address at the previous timestep as can be seen from the interpolation gate, fig.\ \ref{fig:write_head_g}. If we interpret the encoded source sentence for an attentional encoder-decoder as being written to memory, then this is precisely the form of addresses we would see - except that in the case of a NTM the addressing strategy is learned not hardcoded. This suggests that this particular inductive bias built into the attentional encoder-decoder is a sensible one.

\begin{figure}[H]
 
\begin{subfigure}{0.48\textwidth}
\includegraphics[width=0.99\linewidth]{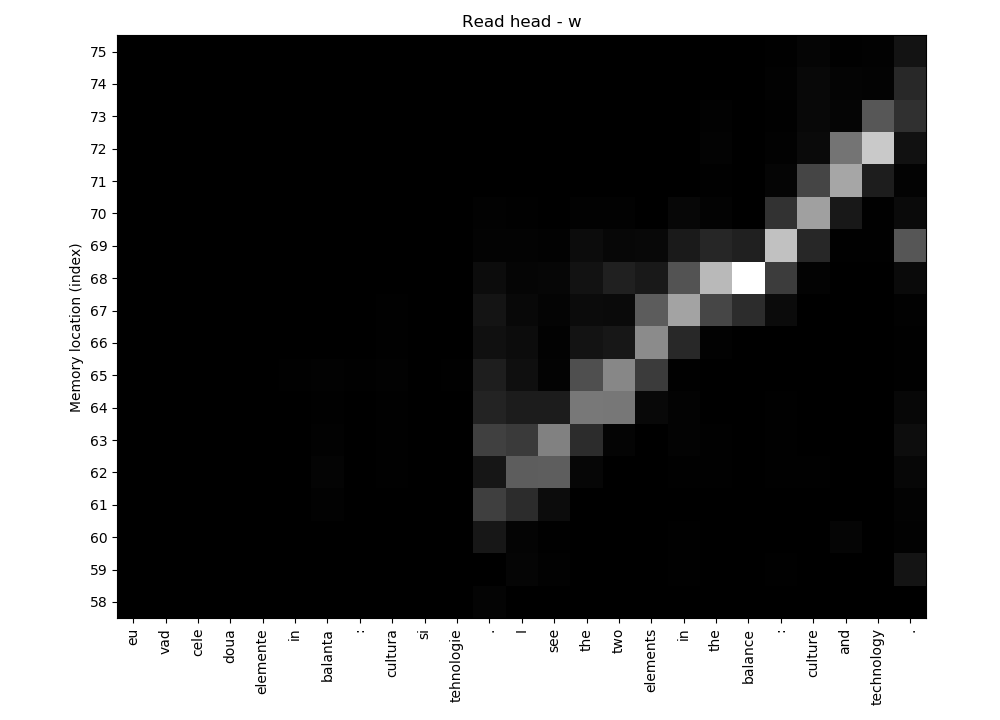}
\caption{Full address}
\label{fig:read_head_w}
\end{subfigure}

\begin{subfigure}{0.48\textwidth}
\includegraphics[width=0.99\linewidth]{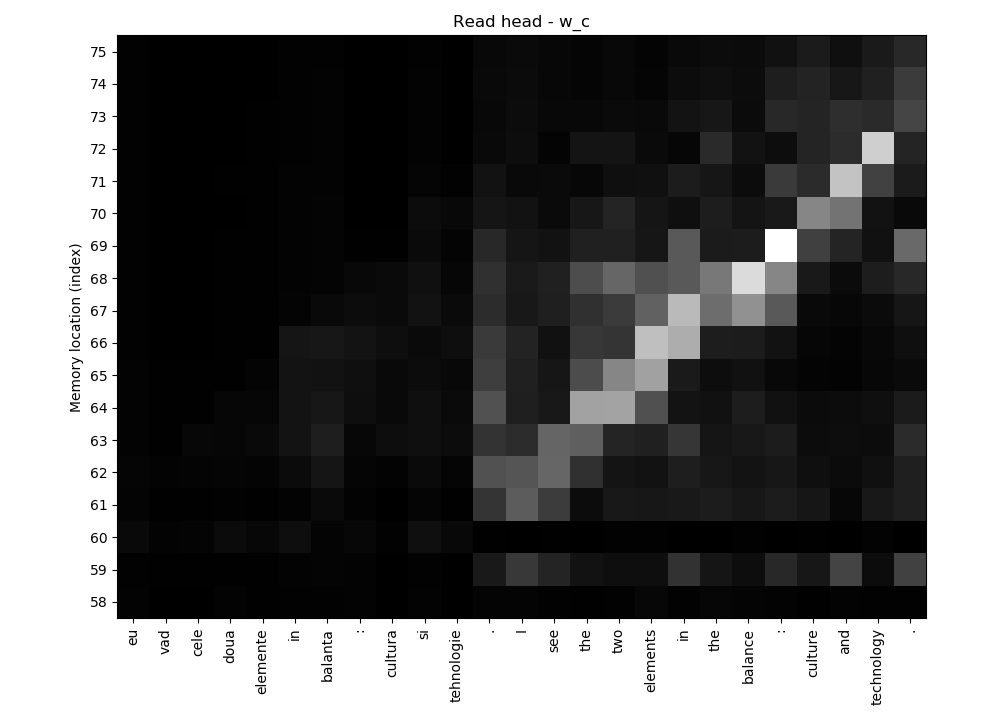}
\caption{Content based addressing}
\label{fig:read_head_w_c}
\end{subfigure}

\begin{subfigure}{0.48\textwidth}
\includegraphics[width=0.99\linewidth]{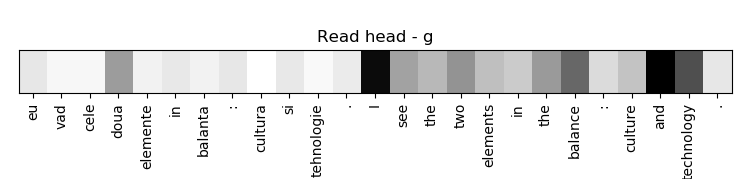}
\caption{Interpolation gate}
\label{fig:read_head_g}
\end{subfigure}

\begin{subfigure}{0.48\textwidth}
\includegraphics[width=0.99\linewidth]{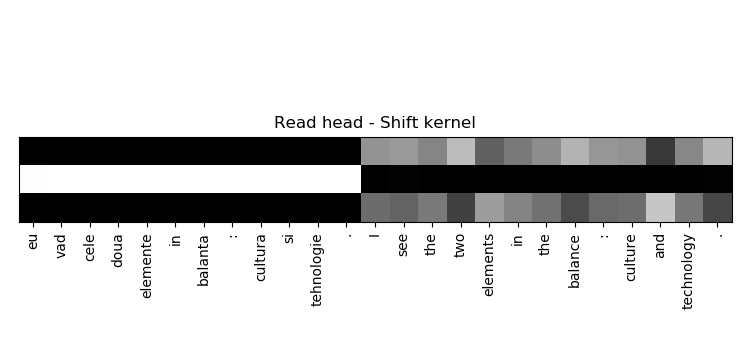}
\caption{Shift kernel}
\label{fig:read_head_s}
\end{subfigure}
 
\caption{Example read head address computation}
\label{fig:read_head}
\end{figure}

The attentional encoder-decoder leaves the encoded source sentence unchanged during decoding as it has no write mechanism. However we observe that the write head is active during decoding for the NTM, fig.\ \ref{fig:write_head_w}. We see that the NTM uses content based addressing, fig.\ \ref{fig:write_head_w_c} to write to the memory locations that are previously read from by the read head , fig.\ \ref{fig:read_head}. This suggests that perhaps the NTM has developed a strategy of marking particular source sentence tokens as completed so as not to retranslate them later during decoding. Interestingly such a mechanism is built directly into the DNC \cite{RN12} and in fact monotonic attention mechanisms have been developed which prevent retranslation of previously translated tokens or preceeding tokens in the source sentence \cite{raffel2017online}. But here of course this strategy is learned from random initialization by the NTM.

Having seen that the NTM learns to write the encoded source sentence to successive memory locations we are not surprised that as the predicted sentence is produced the NTM reads the from memory locations similarly to the attentional encoder-decoder. We see that the previously written to memory locations are then read back from, fig.\ \ref{fig:read_head_w}. Interestingly, we see the read head addresses of the NTM as it produces the predicted sentence are heavily determined by its content based addressing, fig.\ \ref{fig:read_head_w_c}. Thus the NTM does not make significant use of its iteration capability, despite exhibiting the type of monotonic iteration through the source sentence as has been observed with attention mechanisms.

We also note that the read head of the NTM is not particularly focused as the NTM sees the source sentence, fig.\ \ref{fig:read_head_w}. This is somewhat surprising as the results of the read operation are available to the controller at the next timestep and thus could be used to retrieve the encoding of a previous source sentence token or a summary of a section of the source sentence rather than relying on the LSTM controller memory solely for this. We suspect that this behaviour is the result of the read head operation not being available to the write head at the \textit{current} timestep and thus cannot be used to disambiguate the current token as has been the motivation for the successful Transformer NMT model \cite{vaswani2017attention}. Thus, we suggest that extending the NTM and other MANNs depth-wise to have successive rather than parallel operations on the memory matrix at each timestep may be a fruitful avenue of future research.

\section{Conclusion}

We have proposed a series of MANN inspired models for machine translation. Two of these models; NTM Style Attention and the Memory-Augmented Decoder extend the attentional encoder-decoder which has achieved state-of-the-art results on many language pairs. These extensions perform 0.2-0.5 BLEU better than the attentional encoder-decoder alone on the low resource Vietnamese$\rightarrow$English translation task and 0.3-0.9 lower BLEU on the Romanian$\rightarrow$English translation task. We conclude that a content based addressing mechanism is sufficient to encode a strategy of monotonic iteration through source sentences and that enabling the network to express this strategy directly does not significantly improve translation quality. From the Memory-Augmented Decoder results it appears as though extending the memory capacity of the decoder in an attentional encoder-decoder does not offer an advantage, contrary to previous results \cite{RN25}.

Our third proposed model is to just use MANNs directly for machine translation. As far as we are aware we are the first to publish results on MANNs used directly for machine translation. The pure MANN model performs marginally better, +0.2-0.3 BLEU, than the attentional encoder-decoder for the Vietnamese$\rightarrow$English translation task. Performance is 0.3-1.9 BLEU worse for the Romanian$\rightarrow$English translation task. We conclude that MANNs in their current form do not improve over the attentional encoder-decoder for machine translation. Our analysis of the algorithm learned by the pure MANN shows that despite being randomly initialized the pure MANN learns a very similar solution to the attentional encoder-decoder.

We note that the performance gap between the pure MANN and attentional encoder-decoder is not very large and that the pure MANN model is very general and does not incorporate any domain specific knowledge. MANNs are a relatively new architecture that have received less attention than encoder-decoder approaches. We expect that with the development of improved MANN architectures, MANNs could achieve state-of-the-art results for machine translation.

\subsubsection*{Acknowledgements}

This publication emanated from research conducted with the financial support of Science Foundation Ireland (SFI) under Grant Number 13/RC/2106.

\bibliographystyle{mtsummit2019}
\bibliography{mtsummit}

\end{document}